\documentclass[letterpaper]{article} % DO NOT CHANGE THIS
\usepackage[preprint]{aaai2027}  % DO NOT CHANGE THIS # for submission
\usepackage[hyphens]{url}  % DO NOT CHANGE THIS
\usepackage{graphicx} % DO NOT CHANGE THIS
\usepackage{natbib}  % DO NOT CHANGE THIS AND DO NOT ADD ANY OPTIONS TO IT
\usepackage{caption} % DO NOT CHANGE THIS AND DO NOT ADD ANY OPTIONS TO IT
\usepackage{algorithm}
\usepackage{algorithmic}

\usepackage{kotex}
\usepackage{amsmath}
\usepackage{amssymb}

\usepackage{newfloat}
\usepackage{listings}
\DeclareCaptionStyle{ruled}{labelfont=normalfont,labelsep=colon,strut=off} % DO NOT CHANGE THIS
\floatstyle{ruled}
\newfloat{listing}{tb}{lst}{}
\floatname{listing}{Listing}

\usepackage{booktabs}

\title{Flat Score, Amplified Failures:\\
How the Error Budget Masks Damage in Quantized LLM Agents}
\author{
    Jiwon Jang\textsuperscript{\rm 1,2}\equalcontrib,
    Kisu Yang\textsuperscript{\rm 1,3}\equalcontrib,
    Heuiseok Lim\textsuperscript{\rm 3}\corresponding,
    Hyunwoo Park\textsuperscript{\rm 2}\corresponding
}
\affiliations{
    \textsuperscript{\rm 1}VAIV Company\\
    \textsuperscript{\rm 2}Seoul National University\\
    \textsuperscript{\rm 3}Korea University\\
}

\begin{document}

\maketitle

\begin{abstract}
Post-training quantization to 4-bit weights is widely reported to be nearly lossless. We test this claim for multi-turn, tool-calling agents, where it now matters most. On $\tau^2$-bench, across two open-weight model families in dense and MoE variants and two domains (eight cells, 456 episodes each, at 16-, 8-, and 4-bit weights), quantization indeed looks free on the standard metric. No cell shows a score change that survives multiple-comparison correction, and in the cell that carries the largest process damage, equivalence testing bounds the change within $\pm7.5$ points. The process tells a different story. Quantization amplifies the failure the model already exhibits at full precision (tool-name hallucination in telecom, with the same directional trend in retail entity errors) by up to $2.5\times$ in volume ($+17.6$ points per task), while creating essentially no new failures. The failure set is the same at every precision (rank correlation $\geq 0.94$, $0.18\%$ novel events). The score stays flat because the benchmark's ten-error budget absorbs the extra failures. Shrinking the budget to two errors re-exposes a score gap of $17$ points, and it does so only in the one cell where quantization added error volume, exactly as the masking account predicts. A targeted error-repair prompt, run for five telecom models at every precision, removes the damage exactly and only where it lives. Both diagnostics, the per-channel error rate and success under a shrinking budget, come from logs benchmarks already collect; we suggest reporting them alongside task reward.
\end{abstract}
% =====================================================================
% caption and body text reconciled with Results/Setup/Abstract.
% Fixes: (1) correction/noise-floor handling + TOST restored;
% (2) caption defines the figure's pooled quantity so it cannot be
%     confused with Table 1's per-task channel rate;
% (3) ratio metric -> absolute widening (matches Table 2); removed the
%     unverified 0.4x and the five-cell restriction (test spans eight);
% (4) "tools that do not exist" -> outside the agent's tool list;
% (5) contribution 4 aligned with the necessary-not-sufficient finding;
% (6) mitigation forward-connection added (abstract promises it);
% (7) 0.2% -> 0.18%; "explains" -> "is consistent with".
% =====================================================================
\section{Introduction}
 
\begin{figure*}[t]
\centering
\includegraphics[width=\textwidth]{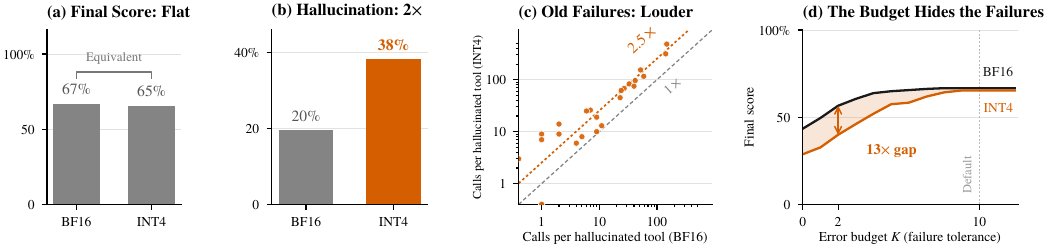}
\caption{\textbf{Flat score, amplified failures.} We run the same agent
benchmark, $\tau^2$-bench telecom with 456 episodes, using a BF16
agent and its AWQ-INT4-quantized version. The cell shown is
Gemma-4-31B. \textbf{(a)} On the benchmark's headline
metric, the final task score, the two agents are statistically
equivalent. Quantization looks free. \textbf{(b)} Underneath, the
share of the agent's tool calls that invoke a tool outside its tool
list doubles, from $19.5\%$ to $38.3\%$. \textbf{(c)} These are not
new failures. Each dot is one hallucinated tool name. Every tool the
BF16 agent already invented is called about $2.5\times$ more often
after quantization, and only $0.18\%$ of the INT4 agent's
hallucinated calls name a tool its BF16 original never used.
\textbf{(d)} The final score stays flat because the benchmark
tolerates ten failed calls per episode and the agent recovers between
failures. Recomputing success under a stricter tolerance $K$
re-exposes the hidden gap, which is 13$\times$ wider at $K{=}2$ than
at the default $K{=}10$.
Models without a pre-existing failure tendency are not amplified at
all, as Table~\ref{tab:main} shows.}
\label{fig:story}
\end{figure*}
 
Weight quantization is the default mechanism for serving large language
models (LLMs) at scale. Reducing weights from 16 to 4 bits cuts memory
and latency several-fold, and systematic evaluations report 4-bit
quantization as nearly lossless, degrading standard benchmark accuracy
by only a point or two~\citep{frantar2023gptq, lin2024awq,
li2024evaluating, jin2024comprehensive}. That verdict, however, rests
almost entirely on \emph{single-turn} evaluations, while deployed LLMs
increasingly act as \emph{agents}, multi-turn tool-calling
controllers that interact with an environment, observe feedback, and
recover from their own mistakes.

\textbf{RQ1:} \emph{Does ``nearly
lossless'' survive the shift to multi-turn tool-calling agents, and
can the benchmark's final score tell?} The natural way to ask turns
out to be misleading.
The obvious experiment is to run quantized and full-precision agents
on an agentic benchmark and compare task success. We do exactly this
on $\tau^2$-bench~\citep{barres2025tau2}, a tool-calling benchmark with
a simulated user, across two open-weight model families
$\times$ \{dense, MoE\} $\times$ two domains (the full
eight-cell grid, 456 simulations per arm), plus two earlier Qwen-3.5
checkpoints in telecom. The result is a flat line
(Figure~\ref{fig:story}a). No score change survives
multiple-comparison correction, and the largest paired decline,
$2.0$ points, sits within $1.3\times$ the run-to-run noise floor.
Taken at face value, quantization is free in the agentic setting too.
Compression studies of agents have so far stopped at this aggregate
verdict~\citep{dong2025acbench}; the rest of the paper asks what it
conceals.
 
We argue this conclusion is an artifact of \emph{what task reward
measures}. Task reward is a binary, end-of-trajectory outcome. It is
swamped by task-difficulty variance and, more importantly, the
agentic loop actively \emph{repairs} the errors quantization
introduces before they reach the outcome. Decomposed at the level of
individual tool calls, trajectories instead reveal a large effect
with two parts. First, universally, \emph{quantization does not
change what the model gets wrong}. In telecom, where the
channel is tool-name hallucination (calls outside the agent's tool
list, nearly always real tools on the \emph{user's} side of the
dual-control interface), the BF16 and INT4 distributions over
hallucinated names are nearly identical, and the invariance holds at
8 bits. The failure set is a property of the \emph{model}, not of
any precision.
Second, in a cell-dependent way, \emph{quantization multiplies how
often that set fires}. In the highest-amplification cell
(Gemma-4 dense, telecom), 4-bit weights produce $2.5\times$ more
out-of-list calls ($649\!\to\!1{,}646$); across cells the multiplier
ranges from ${\approx}1\times$ to $4.6\times$, and in retail the
channel that moves is entity/argument errors instead ($1.45\times$,
Gemma-4 MoE, a trend short of significance). Quantization rescales
the model's existing error set; the scale factor is a property of the
cell.
 
\textbf{RQ2:} \emph{Why does the final score stay flat while the
agent's per-step failures multiply?} Because the loop hides the
damage. $\tau^2$-bench, like other agentic
benchmarks, allows ten failed tool calls per episode, and per-error
recovery is equally reliable at both precisions, so the budget
absorbs the extra failures quantization produces. Tightening the
tolerance from ten errors to two re-exposes the deficit, growing the
INT4 score gap from $1.3$ to $16.7$ points, and it re-exposes damage
only in the one cell where quantization added error volume ($15.4$
points there, at most $0.2$ elsewhere). The score is flat because the
environment is forgiving, not because the model is unharmed. The same
localization predicts the remedy, and a reflexive error-repair
prompt indeed collapses the channel wherever it runs but moves score
only in the damaged cell.
 
\textbf{RQ3:} \emph{Why are some models damaged by quantization
while others are unaffected?} Not size or dense-versus-MoE architecture but
the model's \emph{full-precision failure propensity}. A model that
rarely hallucinates tools at 16 bits (Qwen-3.6) has nothing to
amplify, a model with a standing
tendency (Gemma-4) has that tendency magnified, and an older Qwen-3.5
checkpoint whose gate is slightly open is slightly, but
significantly, amplified. The open gate is necessary rather than
sufficient, since a cell with an equally open gate (Gemma-4 MoE
telecom, $19.3\%$) multiplies by only $1.05\times$. A logit-margin account is
consistent with this pattern. Quantization noise flips only
thin-margin argmax decisions, precisely those the model already got
wrong occasionally, making baseline propensity the observable shadow
of the thin-margin mass.
 
Our contributions are:
\begin{itemize}
\item \textbf{The final score cannot detect 4-bit damage (RQ1).}
End-of-episode task reward is not merely low-powered but
\emph{actively masked} by the environment's error budget; stopping
at task success yields a false ``all-clear.''
\item \textbf{Quantization amplifies the existing failure set
(RQ1).} (i) The failure \emph{set} is precision-invariant across 16,
8, and 4 bits ($0.97$ correlation, $0.18\%$ novel events), ruling
out the ``some models are just worse'' confound; (ii) the failure
\emph{volume} multiplier is cell-dependent, $1\times$ to
$4.6\times$, so damage claims must be made per cell.
\item \textbf{The error budget masks the damage, falsifiably
(RQ2).} A counterfactual budget analysis shows the recovery slack
hides the damage; its advance prediction, that tightening
re-exposes damage only in cells with added error volume, holds
across all cells (one adds volume, nine check the converse),
widening $15.4$
points there versus at most $0.2$ elsewhere; and a reflexive
error-repair system, run for five telecom models at all three
precisions, recovers score only in the damaged cell.
\item \textbf{Full-precision propensity gates the risk (RQ3).}
Neither architecture nor size does. A standing full-precision failure
channel is the \emph{precondition}, and the added error volume the
observable \emph{signature}, of maskable damage. Both come from logs
benchmarks already collect; we suggest reporting them alongside task
reward, and low-slack deployments face more 4-bit risk than
single-turn benchmarks show.
\end{itemize}

\section{Related Work}
 
\paragraph{Post-training quantization and its evaluation.}
Post-training weight quantization~\citep{frantar2023gptq, lin2024awq,
xiao2023smoothquant, dettmers2022llmint8} has become the default
mechanism for serving large language models, and systematic
evaluations consistently report that 4-bit quantization leaves
aggregate benchmark accuracy largely
intact~\citep{li2024evaluating, jin2024comprehensive}. A separate
line of work, however, shows that this aggregate view conceals
non-uniform damage. Compression disproportionately affects long-tail
and underrepresented inputs~\citep{hooker2020compressed,
ahia2021lowresource}, increases hallucination rates in
generation~\citep{li2024dawn, haque2025secure}, and can erode
alignment- and safety-related behaviors~\citep{rath2026quantization}.
These studies establish \emph{that} quantization damage hides below
aggregate metrics, but in single-turn, static generation; none
characterize \emph{where} it goes in a closed-loop agentic
trajectory, nor why end-to-end metrics stay flat while per-step
reliability degrades.
 
\paragraph{Evaluating tool-calling agents.}
Benchmarks for tool use have progressed from single-turn function
calling~\citep{patil2024gorilla}, function-calling
leaderboards~\citep{patil2025bfcl}, and stepwise capability
probes~\citep{chen2024teval, qin2024toolllm} to
long-horizon interactive environments with simulated users, most
notably $\tau$-bench~\citep{yao2024taubench} and its dual-control
successor $\tau^2$-bench~\citep{barres2025tau2}. These environments
score agents by end-of-episode task reward
(pass\textasciicircum{}$k$), and prior analyses note that tool use is
fragile~\citep{cuadron2025saber}, yet the benchmarks ship with
generous per-episode error tolerances allowing retries. All of this work evaluates full-precision models; the
interaction of error tolerance with per-step error rate has not been
examined as a measurement problem.
 
\paragraph{Compression under agentic workloads.}
Closest to our setting, \citet{dong2025acbench} evaluate quantized
and pruned models on agentic capability suites and report
capability-level degradation patterns, and
\citet{paramanayakam2025lessismore} document performance drops when
quantized models serve as on-device function-calling agents. Both
efforts score isolated steps or aggregate accuracy on a fixed
pipeline. We instead evaluate quantized agents closed-loop across
single- and dual-control domains, decompose failures into
domain-specific channels, and trace individual failure events across
precisions within a single model, yielding two findings unavailable
to step-level or aggregate evaluation. Quantization amplifies the model's pre-existing failure
set rather than creating new failures, and the final score is blind to
the amplification because the environment's error budget absorbs the
extra failures until that budget is tightened.
\section{Experimental Setup}

\paragraph{Benchmark.}
We evaluate on $\tau^2$-bench~\citep{barres2025tau2}, a tool-calling
benchmark in which an agent resolves a customer request over a
multi-turn conversation with a simulated user, issuing tool calls
against a stateful environment. We use its two domains.
\textsc{Retail} is a single-control setting in which only the agent
acts on the environment; \textsc{telecom} is a dual-control setting
in which the user also executes actions on their own device and the
agent must reason about state it cannot manipulate directly. The domains differ
in horizon and tool inventory (telecom trajectories are longer,
median $\sim$16 versus $\sim$11 agent turns, with a larger, less
standard tool set) and, as we show, expose different failure
channels. Each domain
contains 114 tasks; an episode ends when the user closes the
conversation, when the agent exhausts \texttt{max\_errors}${=}10$
failed tool calls, or at \texttt{max\_steps}${=}300$. Task reward is the standard
$\tau^2$-bench outcome (a final database-state check plus per-task
assertions). In practice it is binary in both domains, so we report
its mean as the final score in percent.

\paragraph{Models and quantization.}
We study two open-weight LLM families, each in a dense and a
mixture-of-experts (MoE) variant, namely \textbf{Gemma-4-31B} (dense) and
\textbf{Gemma-4-26B-A4B} (MoE, 4B active parameters), and
\textbf{Qwen-3.6-27B} (dense) and \textbf{Qwen-3.6-35B-A3B} (MoE, 3B
active parameters), plus \textbf{Qwen-3.5-27B} and
\textbf{Qwen-3.5-9B} (both dense) as family-version and size
extensions. Each model runs at three precision arms:
\textbf{BF16}, the original 16-bit weights and full-precision
reference; \textbf{FP8}; and \textbf{INT4}, quantized with AWQ. All agents
are served with vLLM on NVIDIA A100 GPUs; the A100 has no native FP8
compute, so the FP8 weights run \emph{weight-only} via the Marlin
kernel (W8A16). Activations and the KV cache are 16-bit in every
arm; all three arms are weight-only. Recipes differ by family
(AWQ-style with $g{=}32$ for Gemma INT4, zero-point AWQ with
$g{=}128$ for Qwen INT4).
The full cross-product (two families $\times$ \{dense, MoE\}
$\times$ two domains, eight cells) is complete at all three
precisions, with both Qwen-3.5 telecom extensions on top. Episodes terminated by \texttt{infrastructure\_error} ($\le 3$
per arm) carry no reward record and are excluded (one retail cell mixed \texttt{max\_steps} of 200
and 300; no episode reaches either cap).

% Table 1 is input here so the table* float lands on the page where
% Results first references it (double-column floats defer to the next
% page top).
% AUTO-GENERATED by workspace/paperfigs.py from raw tau2-bench logs.
% Do not edit numbers by hand -- regenerate.
\begin{table*}[t]
\centering
\small
\setlength{\tabcolsep}{7pt}
\begin{tabular}{ll ccc ccc l}
\toprule
& & \multicolumn{3}{c}{\textbf{Final score (\%)}} & \multicolumn{3}{c}{\textbf{Active-channel error rate (\%)}} & \\
\cmidrule(lr){3-5}\cmidrule(lr){6-8}
Domain & Model & BF16 & INT4 & Change & BF16 & INT4 & $\Delta$ rate & 4-bit verdict\\
\midrule
\textsc{retail} & Gemma-4-31B & 47.5 & 52.1 & n.s. & \;\,1.78 & \;\,2.19 & $+0.41$ & --\\
 & Gemma-4-26B-A4B & 51.8 & 47.6 & n.s. & \;\,4.18 & \;\,5.65 & $+1.48$ & --\\
 & Qwen-3.6-27B & 53.3 & 54.2 & n.s. & \;\,1.70 & \;\,1.58 & $-0.13$ & --\\
 & Qwen-3.6-35B-A3B & 59.9 & 60.4 & n.s. & \;\,1.75 & \;\,1.69 & $-0.06$ & --\\
\midrule
\textsc{telecom} & Gemma-4-31B & 66.7 & 65.4 & n.s. & 19.51 & 38.26 & $\mathbf{+18.75}$\;\, & \textbf{Damaged but masked}\\
 & Gemma-4-26B-A4B & 54.2 & 50.7 & n.s. & 19.26 & 23.05 & $+3.79$ & --\\
 & Qwen-3.6-27B & 96.3 & 96.9 & n.s. & \;\,0.38 & \;\,0.47 & $+0.09$ & --\\
 & Qwen-3.6-35B-A3B & 98.2 & 96.3 & n.s. & \;\,0.04 & \;\,0.26 & $+0.23$ & --\\
 & Qwen-3.5-27B & 98.2 & 96.3 & n.s. & \;\,0.26 & \;\,1.21 & $\mathbf{+0.95}$ & \textbf{Damaged but masked}\\
 & Qwen-3.5-9B & 91.6 & 95.2 & n.s. & \;\,0.61 & \;\,0.43 & $-0.18$ & --\\
\bottomrule
\end{tabular}
\caption{\textbf{Quantization damages two of the ten cells, and
the final score detects neither.} The score change is significant
nowhere (n.s.; seven cells are moreover TOST-equivalent at
$\pm7.5$ points), while the error channel rises significantly in
exactly two cells, whose $\Delta$ rates are \textbf{bold} and whose
verdicts read \emph{damaged but masked}; dashes mark cells whose channel
does not move significantly. Runs are 456 episodes per arm (a few
unscored episodes leave $n{\geq}453$ in three arms). The active channel is entity/argument
errors in retail and tool-name hallucination in telecom; rates are
percentages of the agent's own tool calls. The rate delta is in points, judged by
95\% task-clustered bootstrap intervals (quoted in the text). The identical Qwen-3.6-35B-A3B and Qwen-3.5-27B score rows
are a counting coincidence.}
\label{tab:main}
\end{table*}

\paragraph{Protocol.}
To isolate the effect of \emph{agent} quantization, the simulated user
is held fixed at full precision (Qwen-3.6-27B, BF16) across all
conditions; only the agent's precision varies. In the Qwen-3.6 dense
cells the agent thus shares a base model with the user simulator; the
Qwen-3.5 and Qwen-3.6 MoE cells, which do not, behave identically, so
this does not drive the family contrast. We run \texttt{num\_trials}${=}4$ trials
per task, giving $114\times4=456$ simulations per cell. vLLM serving
is not bit-deterministic (continuous batching perturbs reduction
order); a repeated 456-episode run shifted the mean score by about
$1.5$ points, our coarse single-repeat estimate of the run-to-run
noise floor. Unless noted, comparisons are BF16 versus INT4 within a
fixed model and domain, holding family, architecture, task
distribution, and user simulator constant.

\paragraph{Process metrics and failure channels.}
Beyond task reward, we log every tool call and its outcome. Our primary
process metric is the \emph{channel rate}, the fraction of the agent's
own tool calls (user-simulator calls excluded) that trigger a given
failure channel, pooled over a cell; paired tests use the same rate
computed per task. We decompose failures into two
channels by matching the environment's error string. A
\textbf{tool-name hallucination} is a call to a tool that does not exist
in the domain's inventory (error \texttt{Tool '$X$' not found}); this is
the dominant channel in telecom. An \textbf{entity/argument error} is a
well-formed call whose arguments reference a record that cannot be found
(e.g.\ a wrong customer or order identifier); this is the channel in
retail. A raw ``not found'' count would conflate inventing a tool
with passing a wrong identifier, and the two behave differently. We additionally track trajectory length
(agent turns, tool calls), termination reasons, and, for the mechanism
analysis, the per-task and per-tool-name distributions of hallucinations.

\paragraph{Statistics.}
For each metric we report the change from BF16 to INT4 with a 95\%
confidence interval from $5{,}000$ bootstrap resamples. Because the
four trials of a task share its difficulty, we resample \emph{tasks}
with replacement, keeping all trials of each sampled task (a cluster
bootstrap over the 114 tasks); an interval excluding zero denotes
significance. Episode-level resampling gives narrower intervals; we
report clustered intervals throughout.
Paired-by-task comparisons (removing task-difficulty variance) and
equivalence tests (two one-sided tests, TOST) at $\pm7.5$ points
supplement these where noted. For the mechanism analysis we compute,
within one model$\times$domain, the BF16/INT4 correlation of
hallucinated-name distributions, the fraction of INT4 events unseen at BF16, and the per-task BF16/INT4 rate relationship.

\paragraph{Repair arm.}
For the mitigation study, five telecom models (the grid's four plus
Qwen-3.5-27B) are additionally run at every precision with a
\emph{reflexive repair} loop:
on a failed tool call, the agent is prompted to diagnose the error
message and re-issue a corrected call, up to three attempts. Agent
decoding is greedy in \emph{all} arms (verified from the logs:
identical prefixes always yield identical outputs); repair arms
differ from base only in user-simulator temperature, $0.6$ versus
$0.7$, so we read contrasts against the run-to-run noise floor.

\paragraph{Counterfactual error budget.}
To ask whether the environment's error tolerance hides the damage, we
treat \texttt{max\_errors} as a knob. For a budget $K$, the
\emph{surviving success rate} $S(K)$ is the fraction of episodes that
succeeded using at most $K$ failed tool calls. The projection is
conservative (an episode exceeding $K$ counts as a failure even if it
finished earlier) but applied identically to both
precisions, so the gap $S_{\text{BF16}}(K)-S_{\text{INT4}}(K)$ is a fair
comparison; \texttt{too\_many\_errors} terminations occur at 10--11
counted errors, confirming $K$ matches \texttt{max\_errors}.

\section{Results}

\subsection{Final Task Reward Is Blind to Quantization}
Table~\ref{tab:main} (left) and Figure~\ref{fig:story}a
report the change in the final score from 16- to 4-bit weights. In
all ten cells the 95\% interval includes zero; the largest point
estimate is $4.6$ points and five cells nominally \emph{improve}. The
absence of a difference is not merely low power. In seven cells,
including the highest-signal one below, the change is statistically
\emph{equivalent} to zero by TOST at $\pm7.5$ points, paired by task. In the three remaining Gemma-4 cells
the intervals are wider than the margin, so flatness there means no
detectable change, not equivalence. The largest paired decline, $2.0$ points in Qwen-3.6 MoE telecom at
$p{=}.049$, survives neither multiple-comparison correction nor the
$1.5$-point noise floor and lies well inside the equivalence margin. Paired by task in the highest-signal cell, 4-bit
is worse on 33 tasks, better on 26, tied on 55.
Aggregate or paired, \textbf{task success cannot distinguish a
4-bit agent from a full-precision one}; the rest of this section
shows why that all-clear is wrong.

\subsection{The Damage Is Real, in the Process, and Channel-Specific}
When we decompose trajectories into the two failure channels
(Table~\ref{tab:main}, right, and Figure~\ref{fig:story}b),
a large and systematic effect emerges. In \textsc{telecom},
whose active channel is tool-name hallucination, 4-bit quantization
raises the Gemma-4 dense hallucination rate from $19.5\%$ to $38.3\%$
of the agent's tool calls (${+}18.8$ points, CI $[14.0,23.6]$), a
$2.0\times$ increase, and drives a $9\times$ rise in
error-budget-exhausting terminations ($5\!\to\!46$ of 456 episodes).
Because INT4 trajectories also contain more calls per episode
($7.3\!\to\!9.4$), the $2.0\times$ rate ratio is a $2.5\times$ growth
in event volume ($649\!\to\!1{,}646$), the multiplier we quote. In \textsc{retail}, whose active channel is entity/argument
errors, the same family moves in \emph{that} channel instead
(Gemma-4 MoE, $4.2\%\!\rightarrow\!5.7\%$, ${+}1.5$, CI
$[-0.1,+3.1]$, a trend short of significance under clustered
intervals). Whichever failure the domain makes available,
quantization pushes on it in Gemma-4. The channel structure is
sharp. Retail produces \emph{zero} tool-name events and telecom
entity rates stay ${\le}0.1\%$. Each domain opens exactly one
failure mode, and quantization can only amplify an open channel.

Three observations sharpen the picture. First, Qwen-3.6 is \emph{not}
amplified in any of its four cells. Telecom tool-name stays at
$0.38\%\!\to\!0.47\%$, not significant, with 9 events at BF16,
and retail entity rates are flat. Second, \textbf{architecture is not a stand-alone
predictor}. The one strongly amplified cell is dense, yet the dense
Qwen-3.6 is flat everywhere, and the Gemma-4 MoE shows only
non-significant trends (${+}3.8$ points in telecom, ${+}1.5$ in
retail), so dense-versus-MoE by itself does not sort damaged from
undamaged cells. Third, susceptibility is
\textbf{checkpoint-specific}. The previous-generation Qwen-3.5,
whose gate is slightly open ($0.26\%$), \emph{is} significantly amplified
($\rightarrow 1.21\%$, ${+}0.95$ $[0.5,1.5]$), Gemma-4's pattern
at a twentieth of the magnitude, while its successor and its smaller
9B sibling (gate $0.61\%$) are both flat. The
discriminating variable is the model's own baseline behavior, not
size or architecture (\S~Analysis).

\subsection{The Cliff Is (Mostly) at Four Bits}
Sweeping the intermediate precision shows that 8 bits, weight-only
in our serving setup, is
close to free. No cell degrades significantly in score (one nominally
\emph{improves} by $9.7$ points; with ten cells tested, one
excursion is unsurprising) and no telecom channel moves at FP8; the
jump to $38.3\%$ happens entirely between 8 and 4 bits. One trend
is instructive. Gemma-4 MoE retail already shows its full INT4-sized
entity/argument movement at FP8 (${+}1.40$, CI $[-0.2,3.0]$, not
significant under clustered intervals). If that trend is real, the
safe-precision boundary, too, is a property of the
model$\times$channel pair, not a universal constant.

\begin{figure}[t!]
\centering
\includegraphics[width=\columnwidth]{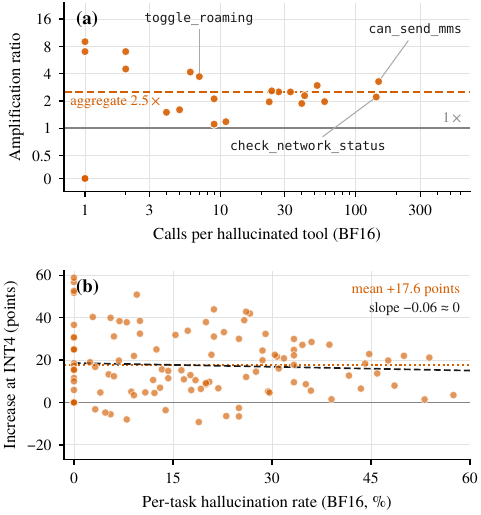}
\caption{Quantization amplifies the existing failure set within a
single model and domain (Gemma-4-31B, telecom). \textbf{(a)} For
each tool name the model invents at BF16, its amplification factor,
INT4 count over BF16 count, versus how often it was already invented.
High-volume tools sit tightly around the aggregate $2.5\times$; the
left-edge scatter is 1--2-event noise. Only 3 of the $1{,}646$ INT4 events
($0.18\%$) name a tool never invented at BF16 (not shown); the BF16/INT4
frequency distributions correlate at $0.97$. \textbf{(b)} Per-task
increase in hallucination rate versus the task's BF16 baseline (114
tasks), a flat cloud around ${+}17.6$ points with least-squares
slope ${-}0.06\approx 0$, so amplification is additive across tasks
rather than proportional to baseline.}
\label{fig:mechanism}
\end{figure}

\subsection{Quantization Amplifies the Existing Failure Set, Not New Ones}
Is the damage a new pathology or a scaling-up of what the model
already did? We answer \emph{within a single model and domain},
Gemma-4 dense telecom, removing any family or architecture confound.

At the level of individual tool names, quantization is almost purely
multiplicative (Figure~\ref{fig:mechanism}a). The set of hallucinated
tool names is essentially unchanged. The per-tool frequency
distributions at BF16 and INT4 correlate at $0.97$, only 3 of the
$1{,}646$ INT4 hallucinations ($0.18\%$) invoke a tool that was never
invented at BF16, and the top-15 fabricated tools at BF16 are all
present at INT4, invoked $2.5\times$ more often
($649\!\rightarrow\!1{,}646$ total).
\textbf{Quantization does not make the model imagine \emph{new}
tools; it makes it reach more often for the ones it was already
tempted by.}

The invented names are not fabrications. Of the 24 distinct
hallucinated names, 23 are real tools \emph{on the user's side} of the
dual-control interface (\texttt{toggle\_airplane\_mode},
\texttt{run\_speed\_test}, \dots; the 24th is a near-miss of the
user-side \texttt{check\_sim\_status}), covering $99.8\%$ of BF16 and
$100\%$ of INT4 events. The channel is \emph{role-boundary
confusion}, the agent reaching for actions only the user can
perform, and quantization amplifies that pre-existing confusion;
retail's structural zero follows, having no user side to reach
across.

At the level of tasks, the amplification takes the form of a
near-uniform shift rather than a proportional scaling
(Figure~\ref{fig:mechanism}b). Regressing the per-task \emph{increase}
on the task's full-precision rate gives a slope of ${-}0.06$,
indistinguishable from zero. The increase is a constant ${+}17.6$
points regardless of where the task started (tercile means $+18.1$,
$+17.2$, $+17.4$); the pooled rise is the slightly larger ${+}18.8$
points because tasks contribute unequal call counts. The propensity signal is therefore strongest as a
\emph{gate}, whether the model opens this failure mode at all
(649 baseline events in Gemma-4 telecom, 9 in Qwen-3.6 telecom, 0
in the retail tool-name channel), rather than as fine per-task
proportionality.
Either way, quantization scales an existing
failure distribution rather than creating a new one.

\begin{figure}[t!]
\centering
\includegraphics[width=\columnwidth]{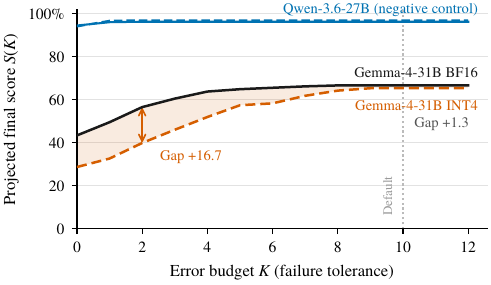}
\caption{The error budget masks the damage. Projected final score
$S(K)$ as the error budget $K$ is tightened, for Gemma-4-31B
telecom and the Qwen-3.6-27B negative control. The BF16--INT4 gap is
$1.3$ points
at the default $K{=}10$ but $16.7$ points at $K{=}2$, roughly
$13\times$. Qwen-3.6, with almost no failures, shows no gap at any
$K$.}
\label{fig:budget}
\end{figure}

\begin{figure}[t!]
\centering
\includegraphics[width=\columnwidth]{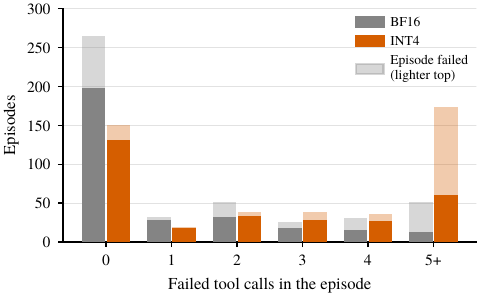}
\caption{Where the extra failures go (Gemma-4-31B, telecom).
Episodes by number of failed tool calls, split into episodes that
still succeeded (solid) and failed (lighter top). Quantization
drains the zero-error bin and swells the ${\ge}5$ tail from 51 to
173 episodes, the region where the 10-error budget starts to bind
(all 46 INT4 budget exhaustions live here). Across the low-error
bins most episodes still succeed. Recovery, not correctness, keeps
the score flat.}
\label{fig:errordist}
\end{figure}

\subsection{Why the Score Stays Flat: the Error Budget Masks It}
Sections above leave a puzzle. If per-step failures rise
$2.5\times$, why is the final score unchanged? Because the agentic loop recovers the
extra failures, and the benchmark's error budget pays for the recovery
(Table~\ref{tab:acct}). Two facts establish this. First, quantization
raises failed tool calls per episode from $1.45$ to $3.64$
($2.5\times$), swelling the ${\ge}5$-failure tail from 51 to 173 of
456 episodes (Figure~\ref{fig:errordist}) and stretching
trajectories ($16.4\!\to\!17.9$ turns, $7.3\!\to\!9.4$ calls;
errors--turns $r{=}0.48$), error-driven recovery that quietly
refunds part of the compute quantization saved. Second, the model recovers
from each individual failure about \emph{equally well} at both
precisions. The recovery rate after at least one failed call is
$55\%$ at both BF16 and INT4. \textbf{Quantization does not
weaken recovery; it simply produces more failures to recover from,
and the $10$-error budget absorbs them.}

Treating that budget as a knob makes the masking explicit and
falsifiable (Figure~\ref{fig:budget}). Recomputing the surviving
success rate $S(K)$ under tighter budgets, the score gap between BF16
and INT4 grows as $K$ shrinks. It is $1.3$ points at the default
$K{=}10$, then $7.5$ at $K{=}5$ and $16.7$ at $K{=}2$, roughly a
$13\times$ widening; the same tightening costs BF16 $10$ points but
INT4 $25$. The Qwen-3.6-27B cell is a negative control. With almost
no baseline failures, masking predicts no hidden gap there, and
indeed it is flat at every $K$, nothing for a tighter budget to
expose.

The budget response doubles as a falsification test. If masking
explains the flat score, tightening should re-expose damage only
where quantization \emph{added} error volume. Across all cells this
is exactly what happens. The gap widens by $15.4$ points in Gemma-4
dense telecom, the one cell with a large added volume, while no other
cell widens by more than $0.2$ points. The largest movement elsewhere
is a $2.2$-point \emph{narrowing} in Gemma-4 MoE telecom, whose event
volume grows only about $5\%$. Where nothing was hidden, nothing is
revealed. The score is flat because the environment is forgiving,
and the damage re-emerges precisely where, and only where, it
exists.
The test is not vacuous, since extra failures landing in episodes
already lost would re-expose nothing, and this partly happens. All
46 INT4 exhaustions score zero yet the default-budget gap is only
$1.3$ points, so part of the damage lands in episodes lost anyway. With one
volume-adding cell the positive arm has a single member; the eight
flat cells check the converse, and a rerun at a tightened budget is
the decisive version of the test.

% Table 2 is input here so the float lands near the mitigation
% discussion that cites it, keeping later pages from being text-only.
% AUTO-GENERATED by workspace/paperfigs.py -- do not edit numbers by hand.
\begin{table}[t]
\centering
\small
\setlength{\tabcolsep}{3.8pt}
\begin{tabular}{l cccc}
\toprule
& \multicolumn{2}{c}{Base} & \multicolumn{2}{c}{With repair}\\
\cmidrule(lr){2-3}\cmidrule(lr){4-5}
Gemma-4-31B, telecom & BF16 & INT4 & BF16 & INT4\\
\midrule
Tool-name hallucination (\%) & 19.51 & 38.26 & \;\,0.42 & \;\,2.90\\
Failed tool calls per episode & \;\,1.45 & \;\,3.64 & \;\,0.07 & \;\,0.21\\
Agent tool calls per episode & \;\,7.29 & \;\,9.43 & \;\,6.31 & \;\,6.20\\
Agent turns per episode & 16.36 & 17.95 & 16.04 & 15.91\\
Budget exhaustions of 456 & 5 & 46 & 0 & 0\\
Recovery rate after a failure (\%) & 55 & 55 & 52 & 52\\
Final score (\%) & 66.67 & 65.35 & 68.86 & 71.27\\
\bottomrule
\end{tabular}
\caption{Recovery accounting in the highest-damage cell. Quantization
creates $2.5\times$ more failures and $29\%$ more tool calls per
episode, yet the recovery rate after a failure is unchanged at
$55\%$. The error budget absorbs the damage until it exhausts,
growing from 5 to 46 episodes. Reflexive repair re-injects the tool
inventory after each failed call, and the channel collapses at both
precisions, exhaustion disappears, and trajectories \emph{shorten}
while the score is preserved.}
\label{tab:acct}
\end{table}

\section{Why Quantization Amplifies Existing Failures}

\paragraph{A logit-margin account.}
The amplify-but-do-not-invent pattern of Figure~\ref{fig:mechanism}
follows from a simple property of quantized decoding. At any generation
step, let the model choose among tokens by their logits, and consider the
correct token $c$ and its strongest competitor $w$, with margin
$m=z_c-z_w$. Weight quantization perturbs the logits by approximately
zero-mean noise $\eta$, so the post-quantization margin is $m'=m+\eta$.
The probability that the argmax flips (that the model now errs)
changes by
$\Delta p(m)=\mathbb{E}_\eta[\sigma(-(m+\eta)/\tau)]-\sigma(-m/\tau)$
for a temperature-like scale $\tau$. It is near zero for large
positive $m$ (confident decisions) and large negative $m$ (already
lost), and \emph{maximized near $m\approx 0$}. Quantization damage
therefore concentrates on thin-margin decisions.

Two predictions follow, both borne out. First, a thin-margin flip
promotes the \emph{already}-closest competitor, the failure the
model occasionally made anyway, so quantization reuses the existing
failure set ($0.97$ correlation, $0.18\%$ novel).
Second, baseline error rate and quantization-induced increase are
governed by the same latent quantity, the mass of thin-margin
decisions, so a model's full-precision failure rate is the
\emph{observable shadow} of its quantization sensitivity; the
prediction is not circular, being tested at a finer grain, the
per-tool-name distribution, than the aggregate rate it predicts.
A positional test separates this account from a natural
alternative in which quantization erodes instruction-following as the
context grows. That account predicts damage concentrated late, but
the INT4-to-BF16 ratio stays within
$1.3\times$--$2.0\times$ in every turn bin and the hallucination rate
\emph{falls} in the longest-context quartile at both precisions.

\paragraph{What governs susceptibility.}
The account explains the family split. Qwen-3.6 adheres almost
perfectly to the tool schema at full precision ($0.38\%$, 9
events). It has essentially no thin-margin mass here and nothing to
amplify, genuinely robust rather than merely saturated, as the
process metric confirms. Gemma-4 carries a standing
tendency to invent plausible tools (649 events), and quantization
magnifies exactly that tendency. The claim is not ``Gemma is worse''
but ``quantization scales up whatever a model is already prone to.''
Architecture does not set this propensity; the checkpoint does. The
Qwen releases have near-zero baselines ($0.26\%$ versus $0.38\%$) yet
land on opposite sides of significance (${+}0.95$ versus ${+}0.09$
points), so near the closed end the gate pins the \emph{ceiling},
damage near $1\%$ versus $38\%$ where it is wide open, rather than
the exact magnitude. The converse is sharper than raw counts
suggest. Gemma-4 MoE telecom opens the gate as wide per call as the
dense variant ($19.3\%$ versus $19.5\%$ at BF16, 841 events) yet
multiplies by only $1.05\times$, per-tool factors scattering
$0.5\times$--$1.5\times$ instead of the dense $2.5\times$. The gate is necessary for
large damage, not sufficient; the aggregate rate cannot see the
thin-margin mass, which varies by cell. The domain contrast follows the same
logic. Telecom's dual-control split places plausible device actions
just across the agent/user role boundary, creating thin-margin
choices between near-synonymous agent- and user-side names;
single-control retail has no boundary and a structurally zero
tool-name channel.

% src/7_mitigation.tex is no longer input: its new findings (15-arm
% repair grid, 71.3% > 66.7% reversal, MoE dissociation, evidentiary
% framing) are folded into the Discussion mitigation paragraph and the
% appendix repair table to fit the 7-page limit. File kept for
% reference.
\section{Discussion}

\paragraph{Agentic benchmarks need process-level reporting.}
End-of-episode reward is blind to quantization for a structural
reason. It is a saturating outcome buffered by the environment's
error tolerance, so a benchmark reporting only task
reward certifies a compressed agent whose per-step reliability has
halved. Benchmarks should report \emph{per-channel error rates}
alongside task reward and, where an error budget exists, success as
a function of that budget;
both come from existing logs and are far more sensitive.

\paragraph{Full-precision propensity is a cheap risk screen.}
Susceptibility is governed by full-precision propensity, so risk can
be screened \emph{before} quantizing by measuring the per-channel
error rate on the target domain. A near-zero rate, as for Qwen-3.6,
signals 4-bit is likely safe there. The
screen is one-directional, an elevated rate flagging risk without
pinning the multiplier, since an equally open gate can stay flat
(Gemma-4 MoE telecom); it still beats size and architecture.

\paragraph{Deployment slack is not always benchmark slack.}
The masking depends on a forgiving environment. Many real
deployments are not forgiving, with single-shot interactions, strict
error limits, irreversible actions, and no silent retries. With the slack gone,
amplified per-step failures become outcome failures, as the
budget-tightening analysis projects. ``Lossless on the benchmark''
need not be lossless where deployed.

\paragraph{A mechanism-targeted mitigation.}
The damage is amplification of a small, \emph{enumerable} set the
model already produces at full precision, with $99.8\%$ of INT4
hallucinations repeating BF16 tool names, so the remedy is
forbidding that set at the source rather than general robustness.
Constrained decoding, native in vLLM, eliminates the dominant
channel by restricting the tool-name field to the inventory. We run the lighter reflexive repair system of Setup for five
telecom models at all precisions; the one significant gain is the damaged
cell, Gemma-4 dense at INT4, ${+}5.9$ points (CI $[+1.1,+10.8]$,
$p{=}.017$) to $71.3\%$, above unmitigated BF16 ($66.7\%$), while
every other interval includes zero. The mechanism is prevention
rather than cure. The in-context inventory collapses the out-of-list
volume from $1{,}646$ events to $82$ and \emph{shortens}
trajectories ($17.9\!\to\!15.9$ turns). Both negative controls behave. Qwen
cells, with nothing to repair, move within or near the
$\pm2.0$-point repeat-run band, and repair trends negative only
where nothing needs fixing (Gemma-4 MoE, whose score dip is
\emph{not} channel-mediated, $-2.6$ at BF16 and $-4.2$ at INT4;
Qwen-3.6 $-2.4$ at FP8; none significant). The same
prompt suppresses its errors only $4\times$ ($882\!\to\!217$)
versus $20\times$ dense, and its trajectories lengthen. Help only where the channel damage lives,
nothing where nothing needs fixing, mild harm elsewhere, a third
confirmation of the localization. The
fix is per-channel. Retail's entity/argument errors instead call for
validate-and-retry or argument grounding, the propensity screen
deciding when either is needed.

\paragraph{Beyond quantization.}
Nothing in the logit-margin account is specific to quantization.
Any weight perturbation adding near-zero-mean logit noise (pruning,
low-rank compression, distillation drift) should likewise
amplify existing thin-margin failures, not create new ones.
The agentic loop accumulates and hides such damage, making
multi-turn tool use a probe for perturbations single-turn accuracy
misses.

\section{Conclusion}
Standard evaluations call 4-bit quantization nearly lossless, and
our ten agentic cells agree, with no final-score change surviving
correction. The agreement is an artifact of what the score can see.
Underneath, quantization amplifies the failures the model already
exhibits, the same tools invented more often, up to $2.5\times$
while creating almost nothing new. The benchmark's error budget absorbs the extra failures, and
tightening it re-exposes the damage exactly where, and only where,
it was added, and a channel-targeted repair recovers the score
there. Susceptibility follows the model's own full-precision
propensity, not its size or architecture. Measure process, not just
outcome.

\bibliography{aaai2027}

@inproceedings{frantar2023gptq,
  title         = {{GPTQ}: Accurate Post-Training Quantization for Generative Pre-trained Transformers},
  author        = {Frantar, Elias and Ashkboos, Saleh and Hoefler, Torsten and Alistarh, Dan},
  booktitle     = {International Conference on Learning Representations},
  year          = {2023},
  eprint        = {2210.17323},
  archiveprefix = {arXiv},
  url           = {https://openreview.net/forum?id=tcbBPnfwxS}
}

@inproceedings{lin2024awq,
  title     = {{AWQ}: Activation-aware Weight Quantization for On-Device {LLM} Compression and Acceleration},
  author    = {Lin, Ji and Tang, Jiaming and Tang, Haotian and Yang, Shang and Chen, Wei-Ming and Wang, Wei-Chen and Xiao, Guangxuan and Dang, Xingyu and Gan, Chuang and Han, Song},
  booktitle = {Proceedings of Machine Learning and Systems},
  volume    = {6},
  pages     = {87--100},
  year      = {2024},
  url       = {https://proceedings.mlsys.org/paper_files/paper/2024/hash/42a452cbafa9dd64e9ba4aa95cc1ef21-Abstract-Conference.html}
}

@inproceedings{xiao2023smoothquant,
  title     = {{SmoothQuant}: Accurate and Efficient Post-Training Quantization for Large Language Models},
  author    = {Xiao, Guangxuan and Lin, Ji and Seznec, Mickael and Wu, Hao and Demouth, Julien and Han, Song},
  booktitle = {Proceedings of the 40th International Conference on Machine Learning},
  pages     = {38087--38099},
  year      = {2023},
  volume    = {202},
  series    = {Proceedings of Machine Learning Research},
  publisher = {PMLR},
  url       = {https://proceedings.mlr.press/v202/xiao23c.html}
}

@inproceedings{dettmers2022llmint8,
  title     = {{LLM.int8()}: 8-bit Matrix Multiplication for Transformers at Scale},
  author    = {Dettmers, Tim and Lewis, Mike and Belkada, Younes and Zettlemoyer, Luke},
  booktitle = {Advances in Neural Information Processing Systems},
  volume    = {35},
  pages     = {30318--30332},
  year      = {2022},
  doi       = {10.52202/068431-2198},
  url       = {https://papers.nips.cc/paper_files/paper/2022/hash/c3ba4962c05c49636d4c6206a97e9c8a-Abstract-Conference.html}
}

@inproceedings{li2024evaluating,
  title     = {Evaluating Quantized Large Language Models},
  author    = {Li, Shiyao and Ning, Xuefei and Wang, Luning and Liu, Tengxuan and Shi, Xiangsheng and Yan, Shengen and Dai, Guohao and Yang, Huazhong and Wang, Yu},
  booktitle = {Proceedings of the 41st International Conference on Machine Learning},
  pages     = {28480--28524},
  year      = {2024},
  volume    = {235},
  series    = {Proceedings of Machine Learning Research},
  publisher = {PMLR},
  url       = {https://proceedings.mlr.press/v235/li24bb.html}
}

@inproceedings{jin2024comprehensive,
  title     = {A Comprehensive Evaluation of Quantization Strategies for Large Language Models},
  author    = {Jin, Renren and Du, Jiangcun and Huang, Wuwei and Liu, Wei and Luan, Jian and Wang, Bin and Xiong, Deyi},
  booktitle = {Findings of the Association for Computational Linguistics: ACL 2024},
  pages     = {12186--12215},
  year      = {2024},
  address   = {Bangkok, Thailand},
  publisher = {Association for Computational Linguistics},
  doi       = {10.18653/v1/2024.findings-acl.726},
  url       = {https://aclanthology.org/2024.findings-acl.726/}
}

@misc{hooker2020compressed,
  title         = {What Do Compressed Deep Neural Networks Forget?},
  author        = {Hooker, Sara and Courville, Aaron and Clark, Gregory and Dauphin, Yann and Frome, Andrea},
  year          = {2019},
  eprint        = {1911.05248},
  archiveprefix = {arXiv},
  doi           = {10.48550/arXiv.1911.05248},
  url           = {https://arxiv.org/abs/1911.05248}
}

@inproceedings{ahia2021lowresource,
  title     = {The Low-Resource Double Bind: An Empirical Study of Pruning for Low-Resource Machine Translation},
  author    = {Ahia, Orevaoghene and Kreutzer, Julia and Hooker, Sara},
  booktitle = {Findings of the Association for Computational Linguistics: EMNLP 2021},
  pages     = {3316--3333},
  year      = {2021},
  address   = {Punta Cana, Dominican Republic},
  publisher = {Association for Computational Linguistics},
  doi       = {10.18653/v1/2021.findings-emnlp.282},
  url       = {https://aclanthology.org/2021.findings-emnlp.282/}
}

@inproceedings{li2024dawn,
  title     = {The Dawn After the Dark: An Empirical Study on Factuality Hallucination in Large Language Models},
  author    = {Li, Junyi and Chen, Jie and Ren, Ruiyang and Cheng, Xiaoxue and Zhao, Xin and Nie, Jian-Yun and Wen, Ji-Rong},
  booktitle = {Proceedings of the 62nd Annual Meeting of the Association for Computational Linguistics (Volume 1: Long Papers)},
  pages     = {10879--10899},
  year      = {2024},
  address   = {Bangkok, Thailand},
  publisher = {Association for Computational Linguistics},
  doi       = {10.18653/v1/2024.acl-long.586},
  url       = {https://aclanthology.org/2024.acl-long.586/}
}

@misc{haque2025secure,
  title         = {Secure or Suspect? Investigating Package Hallucinations of Shell Command in Original and Quantized {LLMs}},
  author        = {Haque, Md Nazmul and Lin, Elizabeth and Arkoh, Lawrence and Tadesse, Biruk and Xu, Bowen},
  year          = {2025},
  eprint        = {2512.08213},
  archiveprefix = {arXiv},
  doi           = {10.48550/arXiv.2512.08213},
  url           = {https://arxiv.org/abs/2512.08213}
}

@misc{rath2026quantization,
  title         = {Quantization Undoes Alignment: Bias Emergence in Compressed {LLMs} Across Models and Precision Levels},
  author        = {Rath, Plawan Kumar and Maliakkal, Rahul},
  year          = {2026},
  eprint        = {2605.15208},
  archiveprefix = {arXiv},
  doi           = {10.48550/arXiv.2605.15208},
  url           = {https://arxiv.org/abs/2605.15208},
  note          = {Accepted at IEEE Cloud Summit 2026; author's accepted version, with the version of record to appear in IEEE Xplore}
}

@inproceedings{patil2024gorilla,
  title     = {Gorilla: Large Language Model Connected with Massive {APIs}},
  author    = {Patil, Shishir G. and Zhang, Tianjun and Wang, Xin and Gonzalez, Joseph E.},
  booktitle = {Advances in Neural Information Processing Systems},
  volume    = {37},
  pages     = {126544--126565},
  year      = {2024},
  doi       = {10.52202/079017-4020},
  url       = {https://papers.nips.cc/paper_files/paper/2024/hash/e4c61f578ff07830f5c37378dd3ecb0d-Abstract-Conference.html}
}

@inproceedings{patil2025bfcl,
  title     = {The Berkeley Function Calling Leaderboard ({BFCL}): From Tool Use to Agentic Evaluation of Large Language Models},
  author    = {Patil, Shishir G and Mao, Huanzhi and Yan, Fanjia and Ji, Charlie Cheng-Jie and Suresh, Vishnu and Stoica, Ion and Gonzalez, Joseph E.},
  booktitle = {Proceedings of the 42nd International Conference on Machine Learning},
  pages     = {48371--48392},
  year      = {2025},
  volume    = {267},
  series    = {Proceedings of Machine Learning Research},
  publisher = {PMLR},
  url       = {https://proceedings.mlr.press/v267/patil25a.html}
}

@inproceedings{chen2024teval,
  title     = {{T}-Eval: Evaluating the Tool Utilization Capability of Large Language Models Step by Step},
  author    = {Chen, Zehui and Du, Weihua and Zhang, Wenwei and Liu, Kuikun and Liu, Jiangning and Zheng, Miao and Zhuo, Jingming and Zhang, Songyang and Lin, Dahua and Chen, Kai and Zhao, Feng},
  booktitle = {Proceedings of the 62nd Annual Meeting of the Association for Computational Linguistics (Volume 1: Long Papers)},
  pages     = {9510--9529},
  year      = {2024},
  address   = {Bangkok, Thailand},
  publisher = {Association for Computational Linguistics},
  doi       = {10.18653/v1/2024.acl-long.515},
  url       = {https://aclanthology.org/2024.acl-long.515/}
}

@inproceedings{qin2024toolllm,
  title         = {{ToolLLM}: Facilitating Large Language Models to Master 16000+ Real-world {APIs}},
  author        = {Qin, Yujia and Liang, Shihao and Ye, Yining and Zhu, Kunlun and Yan, Lan and Lu, Yaxi and Lin, Yankai and Cong, Xin and Tang, Xiangru and Qian, Bill and Zhao, Sihan and Hong, Lauren and Tian, Runchu and Xie, Ruobing and Zhou, Jie and Gerstein, Mark and Li, Dahai and Liu, Zhiyuan and Sun, Maosong},
  booktitle     = {International Conference on Learning Representations},
  year          = {2024},
  eprint        = {2307.16789},
  archiveprefix = {arXiv},
  url           = {https://proceedings.iclr.cc/paper_files/paper/2024/hash/28e50ee5b72e90b50e7196fde8ea260e-Abstract-Conference.html}
}

@inproceedings{yao2024taubench,
  title         = {{$\tau$-bench}: A Benchmark for Tool-Agent-User Interaction in Real-World Domains},
  author        = {Yao, Shunyu and Shinn, Noah and Razavi, Pedram and Narasimhan, Karthik},
  booktitle     = {International Conference on Learning Representations},
  year          = {2025},
  eprint        = {2406.12045},
  archiveprefix = {arXiv},
  url           = {https://proceedings.iclr.cc/paper_files/paper/2025/hash/1b126cc38b8638e07bef37e7b2bb72bf-Abstract-Conference.html}
}

@inproceedings{barres2025tau2,
  title         = {{$\tau^2$-Bench}: Evaluating Conversational Agents in a Dual-Control Environment},
  author        = {Barres, Victor and Dong, Honghua and Ray, Soham and Si, Xujie and Narasimhan, Karthik},
  booktitle     = {Proceedings of the 43rd International Conference on Machine Learning},
  year          = {2026},
  volume        = {306},
  series        = {Proceedings of Machine Learning Research},
  publisher     = {PMLR},
  note          = {Oral},
  eprint        = {2506.07982},
  archiveprefix = {arXiv},
  url           = {https://arxiv.org/abs/2506.07982}
}

@inproceedings{cuadron2025saber,
  title         = {{SABER}: Small Actions, Big Errors -- Safeguarding Mutating Steps in {LLM} Agents},
  author        = {Cuadron, Alejandro and Yu, Pengfei and Liu, Yang and Gupta, Arpit},
  booktitle     = {ICLR 2026 Workshop on Memory for LLM-Based Agentic Systems (MemAgents)},
  year          = {2026},
  eprint        = {2512.07850},
  archiveprefix = {arXiv},
  url           = {https://arxiv.org/abs/2512.07850}
}

@inproceedings{dong2025acbench,
  title     = {Can Compressed {LLMs} Truly Act? An Empirical Evaluation of Agentic Capabilities in {LLM} Compression},
  author    = {Dong, Peijie and Tang, Zhenheng and Liu, Xiang and Li, Lujun and Chu, Xiaowen and Li, Bo},
  booktitle = {Proceedings of the 42nd International Conference on Machine Learning},
  pages     = {14169--14202},
  year      = {2025},
  volume    = {267},
  series    = {Proceedings of Machine Learning Research},
  publisher = {PMLR},
  url       = {https://proceedings.mlr.press/v267/dong25k.html}
}

@inproceedings{paramanayakam2025lessismore,
  title         = {Less is More: Optimizing Function Calling for {LLM} Execution on Edge Devices},
  author        = {Paramanayakam, Varatheepan and Karatzas, Andreas and Anagnostopoulos, Iraklis and Stamoulis, Dimitrios},
  booktitle     = {2025 Design, Automation \& Test in Europe Conference (DATE)},
  pages         = {1--7},
  year          = {2025},
  publisher     = {IEEE},
  eprint        = {2411.15399},
  archiveprefix = {arXiv},
  url           = {https://ieeexplore.ieee.org/document/10992798}
}

% Check whether the conference requires a reproducibility checklist to be included in the paper.
% If so, you can uncomment the following line and adjust the path to include it.
% Appendix for the arXiv/preprint build. Content mirrors the AAAI
% supplementary (supplementary.tex); keep the two in sync. To attach,
% uncomment the \input{appendix} line at the end of main.tex; the
% appendix follows the references, and its floats and labels share the
% paper's numbering (no clashes: appendix labels are tab:repair,
% tab:w8, tab:quant, tab:channel-composition, fig:cliff,
% fig:turncurve, fig:paired-traj, fig:repair-mechanism).
\clearpage
\appendix
\setcounter{secnumdepth}{1}

\section{Additional Results}
\begin{figure}[t]
\centering
\includegraphics[width=\columnwidth]{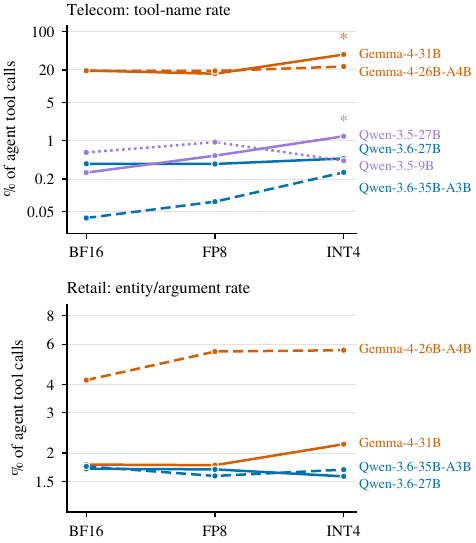}
\caption{The cliff is (mostly) at four bits: the active-channel rate
across the precision sweep BF16$\to$FP8$\to$INT4 for every cell (log
scale; solid dense, dashed MoE, dotted 9B; $*$ marks a significant
change vs.\ BF16). In telecom, rates are flat through FP8 and jump
only at INT4 (Gemma-4-31B, Qwen-3.5-27B). The one FP8 movement is
Gemma-4-26B-A4B retail, whose entity/argument trend is already fully
present at 8 bits, though not significant under clustered intervals.}
\label{fig:cliff}
\end{figure}

\begin{figure}[t]
\centering
\includegraphics[width=\columnwidth]{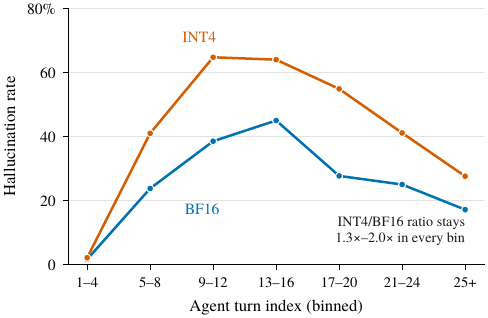}
\caption{Hallucination rate by agent turn index (Gemma-4-31B,
telecom). Both precisions follow the same hump, near zero in the
opening turns, then a mid-trajectory peak and a late decline. The rate
also falls in the longest-context quartile at both precisions (BF16
$34.5\%\to27.2\%$, INT4 $61.8\%\to41.9\%$), and the INT4-to-BF16
ratio stays within $1.3\times$--$2.0\times$ in every bin. An account
in which quantization erodes instruction-following as context grows
predicts the ratio to rise with position; the observed uniformity is
what a per-decision account predicts.}
\label{fig:turncurve}
\end{figure}

Figure~\ref{fig:cliff} visualizes the precision sweep
BF16$\to$FP8$\to$INT4 for every cell of the paper's grid;
Table~\ref{tab:w8} reports the BF16$\rightarrow$FP8 comparison under
the main protocol. Figure~\ref{fig:turncurve} tests whether the
amplification depends on position in the trajectory.
Table~\ref{tab:repair} reports the full repair grid behind the
mitigation paragraph of the Discussion, and a closing
paragraph disentangles three conflated senses of ``horizon.''

\begin{table*}[t]
\centering
\small
\setlength{\tabcolsep}{6pt}
\begin{tabular}{lccc}
\toprule
& \multicolumn{3}{c}{$\Delta$ score, repair $-$ base (points)}\\
\cmidrule(lr){2-4}
Model (telecom) & BF16 & FP8 & INT4\\
\midrule
Gemma-4-31B & $+2.2$ & $0.0$ & $\mathbf{+5.9}$\\
Gemma-4-26B-A4B & $-2.6$ & $+0.4$ & $-4.2$\\
Qwen-3.6-27B & $-0.4$ / $+1.1^{a}$ & $-2.4$ & $+0.9$\\
Qwen-3.6-35B-A3B & $0.0$ & $+0.9$ & $+1.5$\\
Qwen-3.5-27B & $-0.9$ & $+0.7$ & $-2.0$\\
\bottomrule
\end{tabular}
\caption{Effect of the reflexive repair system on the final score
for five telecom models (456 episodes per arm; deltas in points).
Exactly one arm is individually significant under the 95\%
task-paired intervals: \textbf{Gemma-4-31B at INT4} ($+5.9$ points,
CI $[+1.1,+10.8]$, $p{=}.017$), the one cell in which quantization
adds error volume; every other interval includes zero, the widest
being Qwen-3.6-27B at FP8 ($-2.4$, CI $[-5.0,0.0]$) and roughly
$\pm5$ points for Gemma arms. The
localization evidence is this cross-arm pattern rather than a
significant precision$\times$repair interaction, which the sample
does not resolve on its own ($+3.7$ points, CI $[-3.2,+10.7]$). Agent
decoding is greedy in every arm; the repair arms differ from base
only in the user-simulator temperature ($0.6$ versus $0.7$), a
nuisance shared by all fifteen arms, so it cannot produce the
one-cell pattern. $^{a}$Two runs of this configuration under
identical settings; their task-level comparison yields the repeat-run
band of $\pm2.0$ points.}
\label{tab:repair}
\end{table*}

% AUTO-GENERATED by workspace/paperfigs.py -- do not edit numbers by hand.
\begin{table*}[t]
\centering
\small
\setlength{\tabcolsep}{5pt}
\begin{tabular}{ll c c}
\toprule
Domain & Model & $\Delta$ score at FP8 & $\Delta$ channel rate at FP8\\
\midrule
\textsc{retail} & Gemma-4-31B & $\mathbf{+9.7}$ & $-0.01$\\
 & Gemma-4-26B-A4B & $+1.1$ & $+1.40$\\
 & Qwen-3.6-27B & $-1.6$ & $-0.01$\\
 & Qwen-3.6-35B-A3B & $+2.2$ & $-0.16$\\
\midrule
\textsc{telecom} & Gemma-4-31B & $+3.5$ & $-2.51$\\
 & Gemma-4-26B-A4B & $-5.7$ & $-0.15$\\
 & Qwen-3.6-27B & $+1.3$ & $-0.00$\\
 & Qwen-3.6-35B-A3B & $-1.3$ & $+0.04$\\
 & Qwen-3.5-27B & $-1.1$ & $+0.27$\\
 & Qwen-3.5-9B & $+3.5$ & $+0.33$\\
\bottomrule
\end{tabular}
\caption{8-bit quantization, from BF16 to FP8, under the same
protocol and intervals as Table~\ref{tab:main}; deltas are in
points and bold marks an interval excluding zero. No cell degrades
significantly in score and one nominally improves. No channel moves
significantly; the one visible trend is Gemma-4-26B-A4B retail, whose
$+1.40$-point entity movement at 8 bits already matches its
INT4-sized trend.}
\label{tab:w8}
\end{table*}

\paragraph{Horizon is three things, pulling in opposite directions.}
``Longer horizons hurt more'' conflates three quantities that act
oppositely. The \emph{task horizon}, the decisions a task
requires, \emph{generates} damage, since more decisions expose more
thin-margin steps (telecom, median $\sim$16 turns, shows the effect;
retail, $\sim$11, does not). The \emph{budget horizon}
(\texttt{max\_errors}) \emph{hides} it, as the loop recovers the
amplified failures until the tolerance is tightened. \emph{Realized
trajectory length} is neither, being a downstream symptom of
recovery, driven by errors ($r{=}0.48$).

\section{Limitations}
The paper's strongest claims are within-model, robust to the
family confound; four limitations bound the generalization. (i) Two
families plus two further Qwen checkpoints support the propensity
claim as a gate, open versus closed, not a fitted curve; a curve
needs more checkpoints. (ii) The budget-masking result is a
projection from existing trajectories, conservative by construction;
a rerun at tightened budgets would confirm it directly, and the
margin account is likewise indirect, since token log-probabilities
are not in our logs, though Figure~\ref{fig:turncurve} rules out the
leading alternative. (iii) The INT4 recipes differ between families
(Table~\ref{tab:quant}), partly conflating checkpoint with recipe;
the confound is bounded, since the finer-grouped, normally more
accurate recipe belongs to the \emph{amplified} family and the
recipe-matched Qwen-3.5/3.6 pair is separated by baseline propensity
alone, but a same-recipe requantization is the missing control. (iv)
Our evidence spans one benchmark, two domains, and weight-only
recipes on A100 GPUs with a 16-bit KV cache, a fixed full-precision
user simulator, and a telecom-only repair arm; low-bit KV caches,
activation-quantized FP8, other benchmarks, and compressed user
simulators are open.

\section{Reproducibility Details}

\paragraph{Benchmark and protocol.}
All runs use a single $\tau^2$-bench commit with its standard
\textsc{retail} and \textsc{telecom} domains (114 tasks each),
\texttt{num\_trials}${=}4$, seed $300$,
\texttt{max\_steps}${=}300$, \texttt{max\_errors}${=}10$, agent
sampling temperature $0.0$, and user-simulator temperature $0.7$
($0.6$ in the repair arms, which also set an explicit agent seed). One
nonuniformity remains. The Gemma-4-31B retail arms ran at
\texttt{max\_steps}${=}200$ (BF16, INT4) and $300$ (FP8); no episode
in any arm reaches either cap, so the difference is immaterial. The
user simulator is Qwen-3.6-27B at BF16 in every arm, served as a
separate vLLM instance; only the agent model and its precision vary
across cells.

\paragraph{Serving.}
Agents are served with vLLM on NVIDIA A100 GPUs with a 16-bit KV
cache. Because the A100 has no native FP8 compute, vLLM executes the
FP8 checkpoints as weight-only 8-bit via the Marlin kernel (W8A16);
the dynamic per-token activation quantization specified in the FP8
configurations is therefore inactive, and activations are 16-bit in
all three arms.

\paragraph{Quantization configurations.}
Table~\ref{tab:quant} lists each arm's weight-quantization recipe as
read from the served checkpoints' \texttt{quantization\_config}.

\paragraph{Coincidence note.}
The identical score triples of the Qwen-3.6-35B-A3B and Qwen-3.5-27B
telecom rows in Table~\ref{tab:main} are a counting
coincidence. Both cells succeed in 448 of 456 episodes at BF16 and
439 at INT4, while the underlying episode outcomes differ in 16 and
32 episodes respectively.

\paragraph{Analysis.}
Every number in the paper is computed from the raw simulation logs by
a single analysis script (channel regexes, per-cell bootstrap and
paired tests, $S(K)$ projection, and figure/table generation); the
script and per-episode logs will be released upon publication.
\section{Sample Trajectories}

The channel rates in the paper are computed by pattern-matching the
per-episode logs. This section grounds that measurement in raw material.
Figure~\ref{fig:paired-traj} shows one matched pair of episodes, BF16 against
INT4 on the same task; Table~\ref{tab:channel-composition} decomposes the two
active channels into the individual tool names and error strings that produce
the aggregate rates; and Figure~\ref{fig:repair-mechanism} shows what the
reflexive repair loop does to the same failure mode, in the one cell where it
is individually significant.

% =========================== FIGURE: paired trajectory ======================
\begin{table*}[tp]
\centering
\small
\begin{tabular}{lrrr}
\toprule
\multicolumn{4}{l}{\textbf{(a) \textsc{telecom} tool-name channel, Gemma-4-31B}}\\
Tool named by the agent & BF16 & FP8 & INT4 \\
\midrule
\texttt{can\_send\_mms}         & 148 (4.4\%) & 124 (3.8\%) & 484 (11.3\%) \\
\texttt{check\_network\_status} & 142 (4.3\%) & 117 (3.6\%) & 314 (7.3\%) \\
\texttt{check\_apn\_settings}   &  52 (1.6\%) &  49 (1.5\%) & 154 (3.6\%) \\
\texttt{check\_status\_bar}     &  59 (1.8\%) &  35 (1.1\%) & 116 (2.7\%) \\
\texttt{check\_sim\_status}     &  42 (1.3\%) &  44 (1.4\%) &  96 (2.2\%) \\
\texttt{run\_speed\_test}       &  33 (1.0\%) &  31 (1.0\%) &  83 (1.9\%) \\
all further device tools        & 173 (5.2\%) & 154 (4.7\%) & 399 (9.3\%) \\
\midrule
total                     & \textbf{649 (19.5\%)} & \textbf{554 (17.0\%)}
                          & \textbf{1646 (38.3\%)} \\
agent tool calls          & 3326 & 3258 & 4302 \\
\midrule
\multicolumn{4}{l}{\textbf{(b) \textsc{retail} entity/argument channel,
Gemma-4-26B-A4B}}\\
Error returned to the agent & BF16 & FP8 & INT4 \\
\midrule
\texttt{User not found}    & 128 (3.9\%) & 173 (5.1\%) & 191 (5.5\%) \\
\texttt{Order not found}   &   3 (0.1\%) &   7 (0.2\%) &   2 (0.1\%) \\
other unresolved entity    &   5 (0.2\%) &   9 (0.3\%) &   4 (0.1\%) \\
malformed arguments        &   0 (0.0\%) &   0 (0.0\%) &   2 (0.1\%) \\
\midrule
total                      & \textbf{136 (4.2\%)} & \textbf{189 (5.6\%)}
                           & \textbf{199 (5.7\%)} \\
agent tool calls           & 3257 & 3389 & 3485 \\
\bottomrule
\end{tabular}
\caption{Composition of the two active channels, over all 456 episodes per
arm, as counts and as a percentage of agent tool calls. The totals are the
corresponding points of the precision-sweep figure: panel~(a) is the Gemma-4-31B
\textsc{telecom} curve, flat from BF16 to FP8 and jumping at four bits;
panel~(b) is the Gemma-4-26B-A4B \textsc{retail} curve, whose movement is
already complete at eight bits. Two properties of the tool-name channel are
visible only in the decomposition. First, the damage is distributed over the
tool inventory rather than driven by one pathological name: every entry at
least doubles from FP8 to INT4. Second, and more informative, all 22 distinct
names the agent hallucinates across the INT4 arm are \emph{real tools on the
user's side of the interface}, and not one is an invented string. What
quantization degrades here is therefore not name recall but the agent/user
boundary: the model tries to actuate the user's device directly instead of
instructing the user to do so. Panel~(b) counts only unresolved entities and
schema-invalid arguments; refusals that correctly enforce a domain rule
(\texttt{Non-delivered order cannot be exchanged} and similar, 23/35/31 calls)
are excluded, as they are policy behaviour rather than channel errors.}
\label{tab:channel-composition}

\vspace{1.8em}

\begin{minipage}{0.6\textwidth}
\centering
\footnotesize
\setlength{\tabcolsep}{2pt}
\begin{tabular}{l l l}
\toprule
Model & FP8 weights & INT4 weights\\
\midrule
Gemma-4-31B & per-chan., sym. & AWQ-style, $g{=}32$, asym.\\
Gemma-4-26B-A4B & per-chan., sym. & AWQ-style, $g{=}32$, sym.\\
Qwen-3.6-27B & E4M3, $128{\times}128$ & AWQ, $g{=}128$\\
Qwen-3.6-35B-A3B & E4M3, $128{\times}128$ & AWQ, $g{=}128$\\
Qwen-3.5-27B & E4M3, $128{\times}128$ & AWQ, $g{=}128$\\
Qwen-3.5-9B & E4M3, $128{\times}128$ & AWQ, $g{=}128$\\
\bottomrule
\end{tabular}
\captionof{table}{Weight-quantization recipes per arm. Here $g$ is the
quantization group size, $128{\times}128$ denotes block-wise weight
scales, and sym.\ marks symmetric quantization. BF16 is the original
16-bit release in all cases. The Gemma INT4 checkpoints store weights
in the compressed-tensors format with an MSE observer and exclude the
vision tower from quantization, while the Qwen INT4 checkpoints use
standard zero-point AWQ with GEMM kernels. The Qwen-3.5-9B arms use
community releases whose published configurations match these
recipes; their serving-host configurations were not re-read. All FP8 checkpoints are executed
weight-only on our hardware, as described under Serving.}
\label{tab:quant}
\end{minipage}
\end{table*}

\begin{figure*}[tp]
\centering
\small
\begin{minipage}{\textwidth}
\fontsize{9}{9}\selectfont
\hrule\vspace{2pt}
\textbf{(a) INT4} \hfill \texttt{reward 0.0} \quad
\texttt{termination: too\_many\_errors} \quad 43 messages
\vspace{2pt}\hrule
\begin{verbatim}
 1 U   Hi, I'm John Smith. I've been having trouble sending MMS messages from my
       phone [...] My phone number is 555-123-2002.
 2 A   call get_customer_by_phone({"phone_number": "555-123-2002"})
 3 T   {"customer_id": "C1001", "line_ids": ["L1001","L1002","L1003"], [...]}
 4 A   call get_details_by_id({"id": "L1001"})                            [ok]
 6 A   call get_details_by_id({"id": "L1002"})                            [ok]
 8 A   call can_send_mms({})
 9 T   Error: Tool 'can_send_mms' not found.                        <-- error 1
10 A   call can_send_mms({})                             (verbatim repetition)
11 T   Error: Tool 'can_send_mms' not found.                        <-- error 2
       [... two further identical call/error pairs: errors 3, 4 ...]
16 A   "[...] could you please tell me if you are able to make phone calls and
       if your mobile data is working for other things [...]"
17 U   call check_status_bar({}) ; call check_network_status({})   [user side]
19 T   Airplane Mode: OFF [...] Cellular Network Type: 2G [...] Roaming: Yes
23 A   "Since you are on a 2G network, that is likely the cause of the problem.
       [...] please use the `set_network_mode_preference` tool on your phone
       and set the mode to `"4g_5g_preferred"`."           <-- correct diagnosis
24 U   call set_network_mode_preference({"mode": "4g_5g_preferred"})  [applied]
25 T   Preferred Network Mode set to: 4g_5g_preferred | 5G | Excellent
30 U   "[...] the messaging app still says it cannot send MMS messages [...]"
31 A   call check_apn_settings({})
32 T   Error: Tool 'check_apn_settings' not found.                  <-- error 5
       [... five further identical call/error pairs: errors 6-10 ...]
       >>> error budget exhausted (10/10); episode terminated, reward 0.0
\end{verbatim}
\vspace{-4pt}\hrule\vspace{2pt}
\textbf{(b) BF16}, same task and trial \hfill \texttt{reward 1.0} \quad
\texttt{termination: user\_stop} \quad 32 messages
\vspace{2pt}\hrule
\begin{verbatim}
 1 U   Hi, my name is John Smith. [...] unable to send MMS messages [...] I'm
       currently traveling in France [...]
 2 A   "[...] Could you please provide me with your phone number or your
       customer ID? [...]"
 4 A   call get_customer_by_phone({"phone_number": "555-123-2002"})       [ok]
 6 A   call get_details_by_id({"id": "L1001"})                           [ok]
 8 A   call get_details_by_id({"id": "L1002"})                           [ok]
10 A   call enable_roaming({"customer_id": "C1001", "line_id": "L1002"})
11 T   Roaming enabled successfully
12 A   "[...] I noticed that data roaming was disabled for your line [...]
       Could you please try sending an MMS message now [...]"
13 U   call can_send_mms({})                                      [user side]
16 A   "Please use the following tools on your device: 1. `can_send_mms()` to
       confirm [...] 2. `check_network_status()` to see your signal strength
       [...]"                                    <-- delegates, does not call
19 U   call check_network_status({})                              [user side]
20 T   Airplane Mode: OFF [...] Cellular Network Type: 2G [...] Roaming: Yes
22 A   "[...] your phone is currently connected to a 2G network. [...] please
       use the `set_network_mode_preference(mode="4g_5g_preferred")` tool on
       your device."                                      <-- same diagnosis
23 U   call set_network_mode_preference({"mode": "4g_5g_preferred"})  [applied]
28 T   Your messaging app can send MMS messages.
31 U   "[...] Have a wonderful day as well, and safe travels! ###STOP###"
       >>> 0 tool-name errors in the episode; task solved, reward 1.0
\end{verbatim}
\vspace{-4pt}\hrule
\end{minipage}
\caption{One matched pair of episodes, Gemma-4-31B on \textsc{telecom},
task \texttt{[mms\_issue]bad\_network\_preference|\allowbreak user\_abroad\_roaming\_%
disabled\_on\allowbreak[PERSONA:None]}, trial 3; the \texttt{[PERSONA:*]} suffix is part
of the benchmark's task identifiers and persona conditioning is disabled
(\texttt{persona\_config}=\texttt{null}) in every arm.
Columns are message index, speaker
(A\,=\,agent, U\,=\,user, T\,=\,tool result), and content; long content is
elided with \texttt{[...]} and tool results are abbreviated to the fields
under discussion. \emph{The two agents reach the same diagnosis} (2G network,
switch to \texttt{4g\_5g\_preferred}) and the INT4 agent even gets the fix
applied at message~24. Its episode nonetheless scores 0.0, because ten of its
tool calls name device tools that belong to the \emph{user's} side of the
interface (\texttt{can\_send\_mms} and \texttt{check\_apn\_settings}, both
of which the user invokes successfully in the same episode) and exhaust the
\texttt{max\_errors}=10 budget. The failure is located in the tool-name
channel, not in task reasoning, and it is the budget rather than the reasoning
that converts it into a lost point. Note also that the INT4 agent reissues the
identical invalid call rather than revising it, which is the behaviour the
reflexive repair arm targets. Because the user simulator is sampled
stochastically, the two episodes share the task, the trial index and the fixed
BF16 user simulator, but not token-identical user turns; the comparison is
task-level, as everywhere in the paper.}
\label{fig:paired-traj}
\end{figure*}

% =========================== FIGURE: repair mechanism =======================
\begin{figure*}[tp]
\centering
\small
\begin{minipage}{\textwidth}
\fontsize{9}{9}\selectfont
\hrule\vspace{2pt}
\textbf{(a) INT4, base arm} \hfill \texttt{reward 0.0} \quad
\texttt{termination: too\_many\_errors} \quad 60 messages
\vspace{2pt}\hrule
\begin{verbatim}
[38] A  call get_customer_by_phone({"phone_number": "555-123-2002"})     [ok]
[40] A  call check_sim_status({})
[41] T  Error: Tool 'check_sim_status' not found.                  <-- error 1
[42] A  call check_sim_status({})                       (verbatim repetition)
[43] T  Error: Tool 'check_sim_status' not found.                  <-- error 2
        [... eight further identical call/error pairs: errors 3-10 ...]
        >>> error budget exhausted (10/10); episode terminated, reward 0.0
\end{verbatim}
\vspace{-4pt}\hrule\vspace{2pt}
\textbf{(b) INT4, reflexive repair arm}, same task and trial \hfill
\texttt{reward 1.0} \quad \texttt{termination: user\_stop} \quad 52 messages
\vspace{2pt}\hrule
\begin{verbatim}
[34] A  call get_details_by_id({"id": "D1002"})                          [ok]
[36] A  ** repair fired: fail_types=[schema,schema] attempts=2 repaired=yes
        "[...] I'd like to run some diagnostics on your phone to figure out
        why you're still seeing 'No Signal' [...] I will check your device's
        network status and SIM card [...]"
[38] A  ** repair fired: fail_types=[schema] attempts=1 repaired=yes
        "[...] Actually, I'll start by checking your status bar to see
        exactly what icons are visible."
[39] U  call check_status_bar({})                                 [user side]
[40] T  Status Bar: No Signal | Data Disabled | 80%
[42] A  ** repair fired: fail_types=[schema x3] attempts=3 repaired=yes
        "[...] Could you please try removing your SIM card and reinserting
        it (reseating the SIM)? [...]"
[43] U  call reseat_sim_card({})                                  [user side]
[46] A  ** repair fired: fail_types=[schema,schema] attempts=2 repaired=yes
        "[...] I am trying to run some diagnostics on my end [...] but I'm
        encountering some technical difficulties with my tools. [...] I am
        going to transfer you to a human technical support agent [...]"
        call transfer_to_human_agents({"summary": "[...]"})
        >>> 0 tool-name errors in the episode; task solved, reward 1.0
\end{verbatim}
\vspace{-4pt}\hrule\vspace{2pt}
\textbf{(c) What the loop checks, what it injects, and how the arm was
configured}
\vspace{2pt}\hrule
\begin{verbatim}
# Every agent tool call is validated before it reaches the environment:
tool = self._tools_by_name.get(tc.name)
if tool is None:
    errors[i] = f"Unknown tool '{tc.name}'."        # <- tool-name channel
else:
    tool.params.model_validate(tc.arguments)        # <- argument channel

# On failure the loop appends the rejected attempt plus one tool-role
# message per call, then regenerates; up to repair_max = 3 attempts:
"ERROR: {err} Reissue this tool call with corrected, schema-valid
 arguments."

# Arm configuration, as recorded in the run artifact (endpoints masked):
agent llm_args: {"temperature": 0.0, "seed": 300, "mitigation": "repair",
                 "repair_mode": "reflexive", "repair_max": 3,
                 "api_base": "http://<host>:<port>/v1"}
user  llm_args: {"temperature": 0.6,
                 "api_base": "http://<host>:<port>/v1"}
\end{verbatim}
\vspace{-4pt}\hrule
\end{minipage}
\caption{How the reflexive repair loop intercepts the tool-name channel.
Gemma-4-31B INT4 on \textsc{telecom}, the one individually significant cell
of the repair grid, on task \texttt{[service\_issue]airplane\_\allowbreak mode\_\allowbreak
on|\allowbreak break\_\allowbreak apn\_\allowbreak settings|\allowbreak
lock\_\allowbreak sim\_\allowbreak card\_\allowbreak pin|\allowbreak
overdue\_\allowbreak bill\_\allowbreak suspension|\allowbreak
unseat\_\allowbreak sim\_\allowbreak card\allowbreak[PERSONA:None]}, trial 1. In the base arm~(a) the agent names the user-side
tool \texttt{check\_sim\_status} and then reissues the identical call ten
consecutive times, consuming the entire error budget in ten messages. In the
repair arm~(b) the validator in~(c) rejects the same class of call before it
reaches the environment; at each of the four marked turns the regenerated
message is policy-conformant, instructing the user to run the device tool
or escalating, and the episode reaches a reward of 1.0 with no tool-name
errors at all. \emph{Two limitations of this excerpt should be read with it.}
First, the logs retain only the accepted message and its repair record
(\texttt{fail\_types}, \texttt{attempts}, \texttt{repaired}), not the text of
the rejected attempts, so panel~(b) shows where the loop fired and what it
accepted rather than what it discarded; \texttt{fail\_types=schema} is the
validator category that includes \texttt{Unknown tool}, i.e.\ the tool-name
channel. Second, a single episode pair cannot be causal evidence: the base and
repair arms differ in their recorded sampling configuration (see
Reproducibility Details), and the two episodes are not turn-aligned. The
excerpt illustrates the mechanism; the evidence that the mechanism is
localized to this cell is the cross-arm pattern across the repair grid.% REF-UPGRADE
}
\label{fig:repair-mechanism}
\end{figure*}
  % preprint/arXiv build only: appendices after references
% \input{ReproducibilityChecklist.tex}

\end{document}